\pdfoutput=1

\documentclass[letterpaper]{article} 
\usepackage{aaai25}  
\usepackage{times}  
\usepackage{helvet}  
\usepackage{courier}  
\usepackage[hyphens]{url}  
\usepackage{graphicx} 
\usepackage{natbib}  
\usepackage{caption} 
\usepackage{algorithm}
\usepackage{algorithmic}

\usepackage{booktabs}
\usepackage{multirow}
\usepackage{multicol}
\usepackage{soul}
\usepackage{color,xcolor}
\usepackage{enumitem}
\usepackage{graphicx}
\usepackage{dblfloatfix}
\usepackage{amssymb}
\usepackage[utf8]{inputenc}
\usepackage{tcolorbox}
\usepackage{amsmath}

\usepackage{newfloat}
\usepackage{listings}
\DeclareCaptionStyle{ruled}{labelfont=normalfont,labelsep=colon,strut=off} 
\floatstyle{ruled}
\newfloat{listing}{tb}{lst}{}
\floatname{listing}{Listing}
\title{Recording for Eyes, Not Echoing to Ears: Contextualized Spoken-to-Written Conversion of ASR Transcripts}
\author{
    Jiaqing Liu\textsuperscript{\rm 1},
    Chong Deng\textsuperscript{\rm 1},
    Qinglin Zhang\textsuperscript{\rm 1},
    Shilin Zhou\textsuperscript{\rm 2},
    Qian Chen\textsuperscript{\rm 1},
    Hai Yu\textsuperscript{\rm 1},
    Wen Wang\textsuperscript{\rm 1}
}
\affiliations{
    \textsuperscript{\rm 1}Tongyi Lab, Alibaba Group\\
    \textsuperscript{\rm 2}School of Computer Science and Technology, Soochow University\\

    \{mingzhai.ljq,dengchong.d,qinglin.zql,tanqing.cq,yuhai.yu,w.wang\}@alibaba-inc.com
    \{slzhou.cs\}@outlook.com
}

\usepackage{bibentry}

\begin{document}

\maketitle

\begin{abstract}
Automatic Speech Recognition (ASR) transcripts exhibit recognition errors and various spoken language phenomena such as disfluencies, ungrammatical sentences, and incomplete sentences, hence suffering from poor readability. To improve readability, we propose a \textbf{Contextualized Spoken-to-Written conversion (CoS2W)} task to address ASR and grammar errors and also transfer the \textit{informal} text into the \textit{formal} style with content preserved, utilizing contexts and auxiliary information. 
This task naturally matches the in-context learning capabilities of Large Language Models (LLMs). To facilitate comprehensive comparisons of various LLMs, we construct a document-level \textbf{S}poken-to-\textbf{W}ritten conversion of \textbf{A}SR Transcripts \textbf{B}enchmark (\textbf{SWAB}) dataset. Using SWAB, we study the impact of different granularity levels on the CoS2W performance, and propose methods to exploit contexts and auxiliary information to enhance the outputs.
Experimental results reveal that LLMs have the potential to excel in the CoS2W task, particularly in grammaticality and formality, our methods achieve effective understanding of contexts and auxiliary information by LLMs. We further investigate the effectiveness of using LLMs as evaluators and find that LLM evaluators show strong correlations with human evaluations on rankings of faithfulness and formality, which validates the reliability of LLM evaluators for the CoS2W task. 
\end{abstract}

%

\vspace{-4mm}
\section{Introduction}
As can be seen in Figure~\ref{fig:example}, since ASR transcripts aim to provide verbatim transcriptions of oral communications, they often exhibit various spoken language phenomena and informal styles, such as filler words, repetitions, repairs, and fragments, and include ASR errors and ungrammatical text. These characteristics lead to poor readability. To enhance the readability of ASR transcripts, we propose the \textbf{Contextualized Spoken-to-Written conversion (CoS2W)} task, which aims to correct ASR errors and grammatical errors and transfer the informal style to the written and formal style while preserving the content, that is, the CoS2W task is designed to convert verbatim transcripts (``echoing to ears'') into readable documents (``recording for eyes''). Note that different domains vary in their requirements for adjustments on ASR transcripts. Some domains only allow removal of filler words while some domains (e.g., speech analysis for diagnosis) require verbatim transcripts. Nonetheless, the readable documents of CoS2W task are particularly helpful for many domains such as podcasts, education, and meetings (business, project, etc.), where efficiency in information delivery and knowledge acquisition is essential.

Moreover, for a wide variety of downstream tasks on ASR transcripts, such as machine translation, summarization, question-answering~\cite{gupta2021disfl}, and sentiment classification~\cite{DBLP:journals/corr/abs-2308-01776}, we find that even competitive models (including LLMs) perform substantially better on written text than on ASR transcripts, and the performance on the downstream tasks could benefit substantially from applying CoS2W to ASR transcripts. For example, for machine translation of ASR transcripts, we employ the competitive CSANMT model\footnote{\scriptsize\url{https://www.huggingface.co/modelscope-unofficial/damo-csanmt-zh-en-large}} on the development set of the speech translation BSTC dataset~\cite{zhang2021bstc} and find that compared to directly translating ASR transcripts, applying CoS2W (with GPT-4) on ASR transcripts then translating improves the BLEURT~\cite{DBLP:conf/acl/SellamDP20} score from 57.07 to 61.9.

\definecolor{local}{RGB}{222,64,44}
\definecolor{global}{RGB}{246,173,83}

\sethlcolor{yellow}
\begin{figure}[t!]
    \centering
    \includegraphics[width=\linewidth]{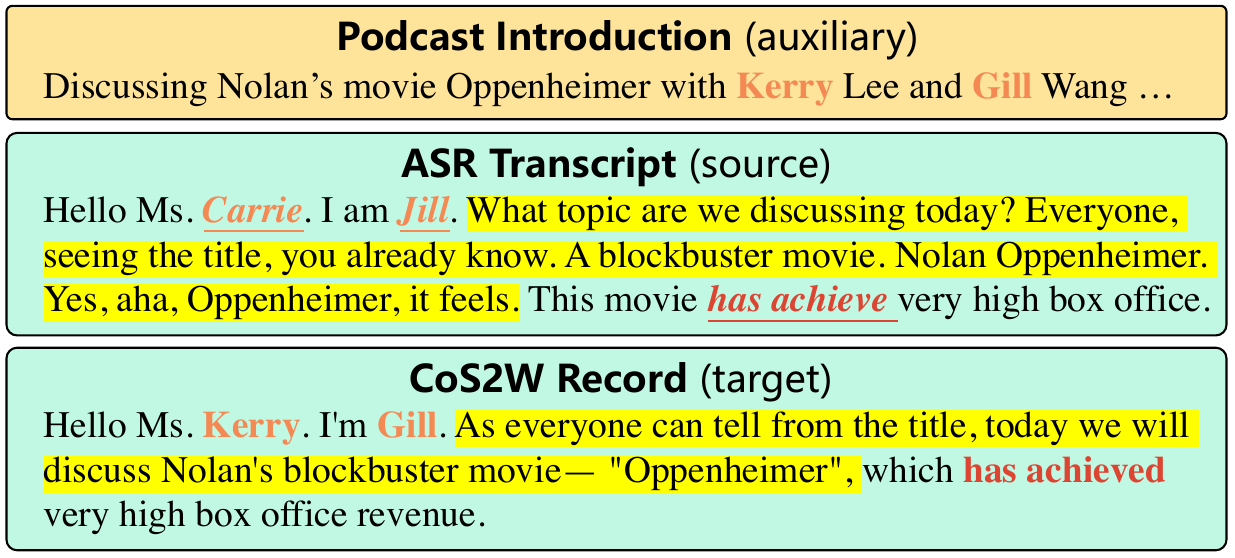}
    \vspace{-6mm}
    \caption{\small{An example of the proposed CoS2W task. CoS2W involves \hl{Text Style Transfer} across sentences, \textcolor{global}{ASR Error Correction} and \textcolor{local}{Grammar Error Correction (GEC)}.}
    }
    \label{fig:example}
    \vspace{-7mm}
\end{figure}

Although the CoS2W task could improve both readability and performance of downstream tasks of ASR transcripts, there is a lack of research efforts on this task due to its significant challenges, including \textit{task complexity}, \textit{effective exploitation of contexts}, \textit{understanding the impact of granularity}, \textit{the challenge in evaluations}, and \textit{scarcity of labeled data}, which we will detail below.

\noindent \textbf{Task complexity.} CoS2W combines many subtasks such as ASR error correction, Grammatical Error Correction (GEC), and text style transfer. Specifically, text style transfer focuses on converting \textit{informal} spoken language expressions (such as repetitions and redundancies) to the formal style.
These subtasks are not simply pipelined but are interconnected: all subtasks are based on a complete semantic understanding of the input and have dependencies among them. ASR and grammar errors can alter semantics, thereby affecting the performance of text style transfer. On the other hand, text style transfer could benefit GEC as GEC performs better on formal texts than spoken texts. 

\noindent \textbf{Understanding the impact of granularity.} Traditional text style transfer focuses on sentence-level conversion. Conversion at a lower granularity (e.g., paragraph-level) instead of sentence-level, may achieve larger improvements to readability, as shown in Figure~\ref{fig:example}. 

However, reducing granularity may introduce faithfulness issues (i.e., divergence from the meaning of the input) as discussed in the results section.

\noindent \textbf{Effective exploitation of contexts.} We define CoS2W as contextualized due to the critical role of contexts in understanding speech transcripts. For example, context is critical to grasp the meanings of sentence fragments in domains with frequent speaker interactions, such as meetings. Some domains also provide auxiliary information for videos and audios, such as introductions of podcasts. 
The auxiliary information could benefit ASR error correction as it may contain relevant named entities, domain terms, and words with challenging pronunciations. This information also facilitates understanding of background knowledge and topics, hence helping the informal-to-formal style transfer. Essentially, the auxiliary information can be considered as \textbf{extended context}. Effectively leveraging auxiliary information to enhance the CoS2W performance poses a challenge.

\noindent \textbf{Challenge in evaluations.} Evaluating the CoS2W task is challenging due to its multifaceted nature, requiring different evaluation methods for its subtasks. Moreover, tasks such as informal-to-formal style transfer remain highly subjective and predominantly depend on human evaluations despite the existence of objective metrics.
LLMs offer a promising opportunity for automatic evaluations. However, leveraging LLMs as reliable evaluators for CoS2W poses many challenges, including designing appropriate prompts, mitigating bias, and ensuring the accuracy, fairness, and reproducibility of the LLM evaluators.

\noindent \textbf{Scarcity of labeled data.} Lack of labeled data for CoS2W hinders both model evaluations and efforts to tackle the challenges discussed above.

The advent of LLMs offers new prospects for addressing the challenges of CoS2W. Firstly, the contextualized CoS2W task matches well with LLMs' in-context learning abilities. Secondly, the CoS2W task considers an output satisfactory as long as it is free of ASR and grammar errors, faithful to the input in semantics, and formal (i.e., close to written text). The generative nature of the problem definition also aligns with LLMs. Thirdly, LLMs have been explored as evaluators for many tasks~\cite{DBLP:journals/corr/abs-2402-11420, DBLP:journals/corr/abs-2304-13462} and achieved notable performance.

Our contributions can be summarized as follows:
\begin{itemize}[leftmargin=*,noitemsep]
    \item We propose the Contextualized Spoken-to-Written conversion (CoS2W) task to improve both readability of ASR transcripts and performance of downstream tasks. To promote research in this field, we construct and make available the \textit{document-level} \textbf{Spoken2Written of ASR transcripts Benchmark (SWAB)} dataset with manual annotations, covering meeting, podcast, and lecture domains in both Chinese and English languages.
    \item We investigate various methods for LLMs to utilize contexts and auxiliary information for CoS2W and provide insights into future research directions.
    \item We find that LLM evaluators show strong correlations with human evaluations on rankings of faithfulness and formality and also analyze their strengths and weaknesses as CoS2W evaluators.
\end{itemize}

\vspace{-2mm}
\section{Related Work}
\label{sec:prior}

\noindent \textbf{ASR Error Correction}
task aims to correct misrecognitions within ASR transcripts. Here we briefly summarize the applications of LLMs and auxiliary information. 

Recent studies~\cite{min2023exploring, yang2023generative} have investigated LLMs' effectiveness in ASR error correction, using varied prompt strategies like few-shot and Chain of Thought (CoT). 
While LLMs show potential in this task, their relatively free generation paradigm often results in unnecessary changes beyond fixing ASR errors, such as paraphrasing error-free text. This behavior affects their performance on metrics like Word Error Rates (WER). 
This problem would also be an issue for the CoS2W task.

In ASR error correction, auxiliary information helps resolve challenging pronunciations, such as N-best transcripts~\cite{ma2023can}, dialogue history~\cite{mai2023enhancing}, video titles and descriptions~\cite{lakomkin2024end}, rare words~\cite{he2023ed}, etc. 
However, this auxiliary information is often brief, consisting of just a few sentences or word lists.
In addition, as far as we know, studies combining auxiliary information with LLMs in this task are relatively scarce.

\noindent \textbf{Grammar Error Correction}
task entails correcting textual grammatical mistakes. Similar to ASR error correction, LLM performance in GEC isn't fully satisfactory~\cite{DBLP:journals/corr/abs-2308-01776, DBLP:journals/corr/abs-2304-01746}. GEC datasets usually follow minimum change principle, clashing with LLMs' relatively free generation paradigm. 
Disregarding objective metrics, human evaluations confirm LLMs' GEC effectiveness~\cite{li2023effectiveness}, which suggests current metrics don't evaluate LLMs fairly and reliably. Therefore, \citet{DBLP:journals/corr/abs-2402-11420} aim to refine GEC evaluation methods, using LLMs to categorize edits and calculate metrics like $F_{0.5}$ to evaluate models. 

Most GEC task corpora are sentence-level, sourced from second-language learners, and categorized into error-coded and direct rewriting paradigms. Table~\ref{tab:dataset} lists common datasets: For English, FCE~\cite{DBLP:conf/acl/YannakoudakisBM11} and AESW~\cite{DBLP:conf/bea/DaudaraviciusBV16} use error-coded, while JFLEG~\cite{DBLP:conf/eacl/TetreaultSN17} and WI-LOCNESS~\cite{DBLP:conf/bea/BryantFAB19} use direct rewriting. Chinese datasets include NLPCC18 and Lang-8~\cite{DBLP:conf/nlpcc/ZhaoJS018}, CGED~\cite{rao2020overview} from HSK essays, and re-annotated YACLC~\cite{DBLP:journals/corr/abs-2112-15043} and MuCGEC datasets~\cite{DBLP:conf/naacl/0004LBLZLHZ22}.

\begin{table*}[t!]
\begin{footnotesize}
    \centering
    \begin{tabular}{lrrccrl}
    \toprule
    \textbf{\fontsize{8pt}{2pt}\selectfont{{Dataset}}} & \textbf{\fontsize{8pt}{2pt}\selectfont{\#Sentences}} & \textbf{\fontsize{8pt}{2pt}\selectfont{Granularity}} & \textbf{\fontsize{8pt}{2pt}\selectfont{Annotation}} & \textbf{\fontsize{8pt}{2pt}\selectfont{Auxiliary}} & \textbf{\fontsize{8pt}{2pt}\selectfont{Language}} & \textbf{\fontsize{8pt}{2pt}\selectfont{Domain}} \\
    
    \midrule
    FCE & 34.0K & sentence-level & \checkmark & $\times$ & EN & FCE Exam Essay \\
    AESW & 1489.2K & document-level & $\times$ & $\times$ & EN & Journal Articles \\
    JFLEG & 1.5K & sentence-level & \checkmark & $\times$ & EN & TOFEL Exam \\
    WI-LOCNESS & 43.1K & sentence-level & \checkmark & $\times$ & EN & Website, Essay \\

    \midrule
    NLPCC18 & 2.0K & sentence-level & \checkmark & $\times$ & CH & Essay \\
    Lang-8 & 717.0K & sentence-level & $\times$ & $\times$ & CH & Language-learning Website \\
    CGED & 7.2K & paragraph-level & \checkmark & $\times$ & CH & HSK Exam \\
    YACLC & 32.1K & sentence-level & \checkmark & $\times$ & CH & Website \\
    MuCGEC & 7.1K & sentence-level & \checkmark & $\times$ & CH & Essay, HSK Exam, Website \\

    \midrule
    Japanese S2W & 18.2K & sentence-level & \checkmark & $\times$ & JA & Conversation, Voicemail \\
    CS2W & 7.2K & sentence-level & \checkmark & $\times$ & CH & Telephone Conversation \\

    \midrule
    SWAB & 29.5K & document-level & \checkmark & \checkmark & EN, CH & Podcasts, Meetings, Lectures \\
    
    \bottomrule
    \end{tabular}
    \vspace{-2mm}
    \caption{Comparison between SWAB and other datasets.}
    \label{tab:dataset}
    \vspace{-6mm}
\end{footnotesize}
\end{table*}

\noindent \textbf{Text Style Transfer}
task aims to modify text to a specific style (e.g., informal to formal, negative to positive) while preserving content. The CoS2W task focuses on informal to formal transfer. 
Recent studies~\cite{DBLP:conf/acl/ReifIYCCW22, DBLP:journals/corr/abs-2401-05707} have employed LLMs to improve text style transfer quality and achieve diverse style conversions. 
In addition, \citet{DBLP:journals/corr/abs-2304-13462} employ the LLM as a multidimensional evaluator and find that it achieves competitive correlations with human evaluation compared to existing automatic metrics, especially in terms of content preservation.

\noindent \textbf{Spoken-to-Written} task
transforms ASR transcripts into formal and readable text, which was initially proposed for Japanese~\cite{DBLP:conf/lrec/IhoriTM20} and has later been extended to Chinese~\cite{DBLP:conf/emnlp/GuoYXJX23}.

Different from previous work, the CoS2W task emphasizes the contextualized ability, aiming to convert whole paragraphs across multiple sentences rather than single sentences, with the help of context and auxiliary information. 
It will further enhance the readability of the results, as illustrated in Figure~\ref{fig:example}. And it aligns well with the in-context learning capabilities of LLMs.
To support this research, we construct the SWAB dataset and report the experiment and evaluation results of LLMs. 
Additionally, our research spans multiple scenarios and languages as shown in Table~\ref{tab:dataset}.

\vspace{-2mm}
\section{The SWAB Dataset}

To conduct experiments and evaluations of the CoS2W task, we construct the \textbf{S}poken2\textbf{W}ritten of \textbf{A}SR Transcripts \textbf{B}enchmark (\textbf{SWAB}) dataset\footnote{\scriptsize\url{https://www.github.com/alibaba-damo-academy/SpokenNLP/tree/main/swab}}, including multiple scenarios (i.e., podcasts, meetings, lectures) in both Chinese and English.
There are 60 transcripts with auxiliary information, with each subcategory comprising 10 documents. 
It is important to note that the SWAB is \textbf{document-level} with complete annotations for entire transcripts, providing an annotated target as a reference for each paragraph.
Furthermore, we provide links to the original audio or video along with timestamps to support multi-modal research.

\vspace{-2mm}
\subsection{Data Source}

We collect Chinese and English data to compare the multilingual capabilities of LLMs, by leveraging the online resources available and selected open-source corpora of spoken transcripts.
Additionally, we select three typical scenarios (i.e., meeting, podcast, and lecture) where efficiency in information delivery and knowledge acquisition is essential.
Among the three scenarios, meetings are high-frequency interactive discussions among multiple participants; podcasts typically involve chats and interviews between two or more individuals; and lectures are monologues from a single person. Essentially, these three domains exhibit significant differences in interactivity, number of participants, length, etc, hence differing from written texts to varying extents and in turn posing diverse levels of challenges to the CoS2W task.

\noindent \textbf{Meetings} are sourced from an open-source meeting corpus with manually annotated information as auxiliary information. 
Chinese meetings are sampled from the training dataset of the AliMeeting corpus~\cite{Yu2022M2MeT}. We select the title and manually annotated topic titles (i.e., sub-topics) as auxiliary information.
English Meetings are sampled from the AMI corpus~\cite{carletta2005ami}. We choose third-party annotated abstractive summarization as auxiliary information, which includes sections on decisions, action items, etc.

\noindent \textbf{Podcasts} are derived from many Chinese and English podcast programs covering a multitude of topics on YouTube\footnote{\url{https://www.youtube.com}}. We collect the podcast introduction as auxiliary information, which provides rich background knowledge such as guest names, specialized terminology, and discussion topics.

\noindent \textbf{Lectures} are sourced from many Chinese and English individual speeches on YouTube, covering a variety of topics. Both Chinese and English lectures are provided with meta-information of the speeches as auxiliary information, which includes the name and biography of the speaker, as well as the title, category, and description of the video.

\vspace{-2mm}
\subsection{Dataset Construction}

We design a relatively free annotation paradigm like text style transfer and direct rewriting of GEC. Targets are considered ``correct'' as long as they fix ASR and grammatical errors, and adopt a written and formal style while faithfully preserving the original content. 
In addition, we focus on maintaining consistency of paragraph boundaries between source and target of CoS2W, which facilitates flexible division of content across the document and fair comparison at different granularity levels.

All collected data are first transcribed using a competitive ASR system~\cite{gao2022paraformer}\footnote{\url{https://www.github.com/modelscope/FunASR}}, then the ASR 1-bests are processed with punctuation insertion,  paragraph segmentation~\cite{zhang2021sequence}, and speaker diarization~\cite{zheng20233d}. 
This ASR structure is widely utilized and operates at peak performance while preserving the overall generality of the dataset.
We utilize GPT-4 (gpt-4-0125-preview)\footnote{\url{https://platform.openai.com/docs/models}} to obtain the initial target of CoS2W for the SWAB dataset. The input is chunk-level to leverage local context and to reduce token consumption. The auxiliary information is also provided to enhance the ASR error correction performance. The prompt for this procedure is in the appendix.

The results of GPT-4 still contain some issues (as shown in our analysis). Consequently, we use human annotation for manual revisions. 
We recruit more than ten college-educated Chinese annotators. Each annotator must undergo comprehensive training to understand the annotation guidelines and meet the accuracy standards on test data before they can officially begin annotating. 
Given the transcription (source), model results (target), and auxiliary information, each qualified annotator must thoroughly review and annotate the entire transcript from beginning to end, to ensure consistency at the document level. 
They are responsible for (1) ensuring paragraph-level consistency between source and target, (2) correcting any remaining ASR errors and grammatical errors, and (3) guaranteeing that the content preserves the original content and embodies a written and formal style. 
During the annotation process, any confusing cases will be discussed, and the annotation guidelines and examples will be continuously updated to ensure clarity and accuracy.

Quality assurance is managed by three senior annotators. 
They will conduct subjective quality checks on 10\% to 20\% of the sampled paragraphs and utilize heuristic methods to perform rule-based inspections on all paragraphs. 
For documents with an error rate of less than 10\%, identified and similar mistakes must be corrected. For documents with an error rate greater than 10\%, a complete revision is required until the document passes the inspection.

We observe that the CoS2W task shows significant diversity. Due to labeling costs, we currently provide only one target but aim to expand target diversity in future work.
Therefore, some objective metrics may not adequately reflect the performance. As a result, we have adopted metrics that focus more on semantics and introduced evaluations by LLMs and humans to mitigate the limitations of a single target.

\vspace{-2mm}
\begin{figure}[ht]
    \centering
    \includegraphics[width=\linewidth]{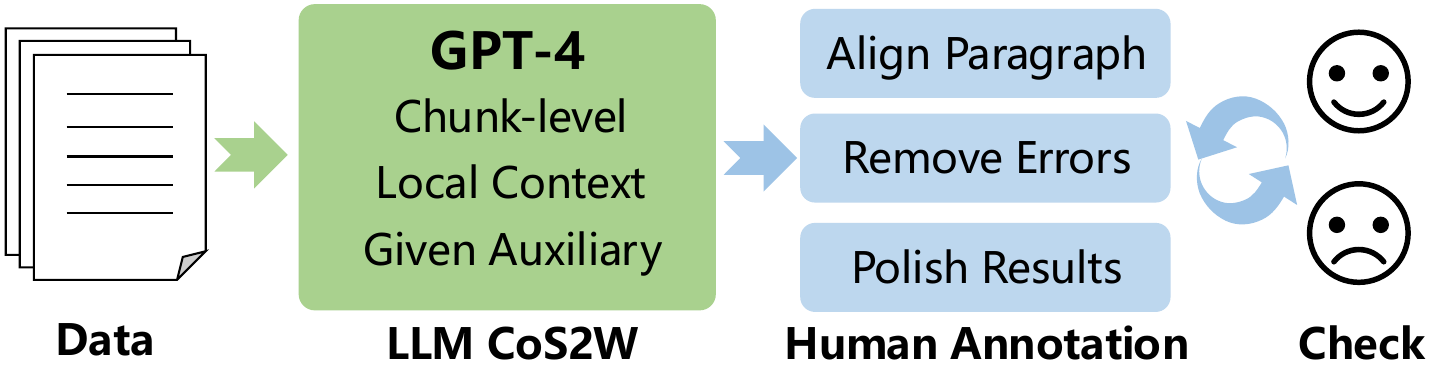}
    \vspace{-6mm}
    \caption{\small{The pipeline of constructing the SWAB dataset.}
    }
    \label{fig:pipeline}
    \vspace{-7mm}
\end{figure}

\subsection{Dataset Analysis}
The target length of the final results is streamlined to 85.40\% of the source length. 
Additionally, annotators adjust 10.43\% of the paragraph segmentation results to improve segmentation and ensure paragraph-by-paragraph consistency between the source and target.
In total, 20.09\% of the paragraphs are identified and corrected for ASR errors. Within the GPT-4 results, 11.29\% of the paragraphs still contain ASR errors and are further modified manually by annotators. 
Only 2\% of the paragraphs in the GPT-4 results have grammatical errors to be corrected manually.
Annotators further optimize 20.26\% paragraphs of GPT-4 to achieve a better written and formal style. Note that further optimization does not imply errors in the formal style of GPT-4 results.

\vspace{-3mm}
\section{Method}

\begin{table*}[t!]
    \centering
    \begin{tabular}{llcccccc}
    \toprule
    \textbf{Model} & \textbf{Auxiliary} & \textbf{BLEU$\uparrow$} & \textbf{ROUGE-L$\uparrow$} & \textbf{BLEURT$\uparrow$} & \textbf{CK-Recall$\uparrow$} & \textbf{S-Faithful$\uparrow$} & \textbf{S-Formal$\uparrow$} \\
    
    \midrule
    Source & None & 16.22 & 41.51 & 55.19 & 12.89 & 7.53 & 4.97 \\
    \midrule
    LLaMA3-8B & None & \phantom{0}8.94 & 20.73 & 40.53 & 13.18 & 6.35 & 7.33 \\
    QWen-14B & None & \phantom{0}3.41 & 15.96 & 30.14 & 11.46 & 3.72 & 6.04 \\
    QWen-Max & None & \phantom{0}9.47 & 28.92 & 50.58 & 22.35 & 6.00 & 7.55 \\
    GPT-4 & None & \textbf{15.51} & \textbf{39.66} & \textbf{62.15} & \textbf{22.49} & \textbf{7.32} & \textbf{8.20} \\

    \midrule
    \multirow{4}{*}{GPT-4} & + Origin & 15.33 & 39.53 & 62.73 & 43.70 & 7.38 & 8.09 \\
     & + RAG & 15.22 & 39.56 & 62.48 & 37.11 & 7.39 & 8.07 \\
     & + Summary & \textbf{15.51} & \textbf{39.98} & \textbf{62.75} & 40.40 & 7.43 & 8.11 \\
     & + Keywords & 15.08 & 39.34 & 62.72 & \textbf{45.42} & \textbf{7.46} & \textbf{8.16} \\
    
    \bottomrule
    \end{tabular}
    \vspace{-2mm}
    \caption{\small{The performance comparison between closed-source LLMs and open-source LLMs at the chunk-level granularity, with different auxiliary information utilization strategies, including directly providing (Origin), retrieving the top-10 most relevant sentences (RAG), based on a summary (Summary), or a list of keywords (Keywords) derived from LLMs. Source refers to the original ASR transcripts.}}
    \label{tab:baseline}
    \vspace{-2mm}
\end{table*}

\begin{table*}[t!]
    \centering
    \begin{tabular}{llcccccccc}
    \toprule
    \multirow{2}{*}{\textbf{Granularity}} & \multirow{2}{*}{\textbf{Context}} & \multirow{2}{*}{\textbf{BLEU$\uparrow$}} & \multirow{2}{*}{\textbf{ROUGE-L$\uparrow$}} & \multirow{2}{*}{\textbf{BLEURT$\uparrow$}} & \multirow{2}{*}{\textbf{CK-Recall$\uparrow$}} & \multicolumn{2}{c}{\textbf{S-Faithful$\uparrow$}} & \multicolumn{2}{c}{\textbf{S-Formal$\uparrow$}} \\
     &  &  &  &  &  & Para. & Chunk & Para. & Chunk \\
    
    \midrule
    Document & None & \phantom{0}1.80 & \phantom{0}4.97 & 10.64 & \phantom{0}5.30 & 2.28 & 2.44 & 4.57 & 3.74 \\
    Chunk & None & 15.51 & \textbf{39.66} & 62.15 & \textbf{22.49} & 7.32 & \textbf{7.57} & 8.20 & \textbf{7.81} \\
    Paragraph & None & \textbf{16.32} & 36.74 & \textbf{63.32} & 19.20 & \textbf{7.99} & 6.60 & \textbf{8.31} & 7.55 \\

    \midrule
    \multirow{3}{*}{Paragraph} & + Local & \textbf{16.44} & \textbf{39.73} & \textbf{67.11} & 22.21 & \textbf{8.19} & 7.41 & \textbf{8.40} & 7.76 \\
     & + Global & 15.55 & 38.73 & 66.00 & 23.64 & 7.89 & 7.05 & 8.36 & 7.74 \\
     & + Both & 16.38 & 39.60 & 67.10 & \textbf{23.78} & 8.18 & \textbf{7.42} & 8.39 & \textbf{7.80} \\
    
    \bottomrule
    \end{tabular}
    \vspace{-2mm}
    \caption{\small{The performance of GPT-4 at different levels of granularity including document, chunk, and paragraph under different contexts. Additionally, S-Faithful and S-Formal show results for both paragraph-level and chunk-level evaluations.}}
    \label{tab:context}
    \vspace{-6mm}
\end{table*}

We compare the performance on the CoS2W task of different LLM across various granularity levels, and introduce contexts and auxiliary information to enhance model performance. 
To evaluate comprehensively, we employ automatic metrics, the LLM Evaluator, and human evaluation.

\vspace{-2mm}
\subsection{Different LLMs}
To assess different LLMs' abilities in solving this complex task, we choose 4 typical LLMs across both open-source and closed-source models. 
The open-source LLMs selected are QWen-14B\footnote{\url{https://www.modelscope.cn/models/qwen/Qwen-14B-Chat}} and LLaMA3-8B\footnote{\url{https://www.github.com/meta-llama/llama}}, chosen for their suitable computational demands for future SFT-related work. For closed-source LLMs, we pick GPT-4 (gpt-4-0125-preview)\footnote{\url{https://platform.openai.com/docs/models}} and QWen-Max (qwen-max-0107)\footnote{\scriptsize\url{https://www.alibabacloud.com/help/en/model-studio/getting-started/models}}, to obtain the state-of-the-art performance of LLMs on this task.

\vspace{-2mm}
\subsection{Different Granularity Levels}
To explore the impact of granularity level, we construct comparisons at the document level, chunk level, and paragraph level, ranging from coarse to fine granularity. 
Document-level granularity requires the model to rewrite all paragraphs of one entire document, using paragraph indexes to correlate the source and target. 
At the chunk-level granularity, we divide the document into chunks based on length (1.5K tokens). The model is required to rewrite each paragraph within the chunk on a paragraph-by-paragraph basis.
For paragraph-level granularity, we conduct experiments focusing on the revision of individual paragraph content. 
Furthermore, we set chunk-level granularity as the baseline, as it allows the use of local context while requiring fewer tokens.

\vspace{-2mm}
\subsection{Context}
Context is essential in text understanding, especially for spoken ASR transcripts. We divide context into local context (neighboring paragraphs) and global context (semantically related but non-adjacent paragraphs)~\cite{han2021modeling}. The global context is acquired by retrieving top-k relevant paragraphs using BM25~\cite{DBLP:journals/ftir/RobertsonZ09}.
For chunk-level experiments, paragraphs within a chunk naturally possess local context. 
To assess the influence of context more accurately, we conduct paragraph-level experiments. By providing 4 local or global paragraphs, we compare the effects of context.

We utilize auxiliary information to enhance ASR error correction by one-step and two-stage methods. 
The one-step method integrates full or retrieved auxiliary information as a reference part of the context in the prompt. The retrieved part is top-K relevant sentences of auxiliary information for current content based on BM25. 
For the two-stage approach, we first use LLMs to extract the summary or keywords from the auxiliary information, with the prompts detailed in the appendix. 
We require the summary to retain all key terms as much as possible, especially challenging words for ASR. 
The keywords should be categorized by their entity types. 
Next, we use either the summary or keywords as references. 

\vspace{-2mm}
\subsection{Evaluations}
\label{ssec:evaluation}
We thoroughly evaluate using objective metrics, LLM Evaluator, and human evaluation.
For objective metrics, we calculate BLEU~\cite{DBLP:conf/acl/PapineniRWZ02}, ROUGE~\cite{lin2004rouge}, and BLEURT~\cite{DBLP:conf/acl/SellamDP20} with the human target as reference.  
It's important to emphasize that the capabilities of BLEU and ROUGE are limited by only one target of the SWAB dataset. Therefore, we focus more on semantic-oriented automatic metrics such as BLEURT. 

To measure the performance of ASR Error Correction, we design the Challenging Keyword Recall (CK-Recall) metric. The challenging keywords are a subset of ASR errors in transcripts, focusing specifically on named entities (e.g., person names, podcast titles). Correcting these errors can significantly improve the reading experience. Furthermore, these words typically cannot be modified or adjusted, making it convenient to directly search in the results. We use the recall calculation formula, where the numerator represents the number of keywords that appear in the results, and the denominator represents the total number of keywords.

In addition, we utilize an LLM (i.e., GPT-4) as the evaluator with the prompts shown in the appendix. We request the LLM to evaluate and score the model-generated target across faithfulness (S-Faithful) and formality (S-Formal). Scores range from 1 (worst) to 10 (best) for each criterion.
S-Faithful reflects content preservation to retain original meaning, and S-Formal reflects the degree of formality in style.
For chunk-level results, we require the outputs to align each paragraph by the paragraph index, so we evaluate them at the paragraph-level granularity. 
To fairly compare paragraph-level and chunk-level results, we evaluate them both at paragraph and chunk levels. 
To enhance effectiveness, we sample a maximum of 100 paragraphs per document for evaluation. 
Furthermore, we sample over 200 paragraphs, whose results of different LLMs are ranked by human evaluators.

\vspace{-2mm}
\section{Results}

\subsection{Results of Different LLMs}
\label{ssec:result_llms}

As shown in Table~\ref{tab:baseline}, we compare GPT-4 and QWen-Max (closed-source LLMs) with LLaMA3-8B and QWen-14B (open-source LLMs) at the chunk-level granularity. 
GPT-4 outperforms all other models across all metrics, followed by QWen-Max. Due to the limitations in model size, the performance of open-source models isn't as good. Among them, the LLaMA3-8B model exhibits superior performance compared to the QWen-14B model.

\textbf{LLMs show potential in solving this complex task, yet still encounter ``faithfulness problems''.} 
The performance of LLMs on the faithfulness score is unsatisfactory, even for state-of-the-art models like GPT-4 (7.32). 
We observe some ``paragraph drift'' situations at the chunk level. For example, the end of one paragraph might be migrated to the beginning of the following one. Even worse, the target and source cannot be aligned by paragraph index. 
Furthermore, faithfulness problems occur even when alignments are correct. 
The model frequently prioritizes coherence and fluency over faithfulness, such as fabrications for fragments or the omission of disorganized insertions in spoken scenarios, which compromises the accuracy of some essential information. 

\vspace{-2mm}
\subsection{Results of Granularity Levels and Contexts}
\label{ssec:context}

As presented in Table~\ref{tab:context}, we compare the performance at different levels of granularity based on the GPT-4 model.
Although the SWAB dataset is constructed at the document level, the CoS2W task at the \textbf{document level still presents a challenge}. It requires a considerable number of tokens for both input and output (about 8K), which is difficult even for GPT-4 as shown in Table~\ref{tab:mismatch}. For document-level results, 76.10\% of the paragraphs are blank without the corresponding index, compared to 1.60\% at the chunk-level granularity. And there is a noticeable decline in all metrics.

Due to ``paragraph drift'' issues at the chunk level as mentioned before, we present LLM evaluation results (i.e., S-Faithful and S-Formal) for both paragraph and chunk levels to ensure a fair comparison. The chunk-level evaluations show that the CoS2W results perform better at chunk-level (7.57 and 7.81) than at paragraph-level (6.60 and 7.55) granularity. This highlights the \textbf{advantages of utilizing local context at the chunk level}. Chunk-level granularity enables LLMs to naturally leverage the local contexts to enhance faithfulness and formality performance. 

In addition, we experiment with various contexts at paragraph level explicitly, \textbf{confirming the role of context in enhancing text understanding}, as is shown in Table~\ref{tab:context}. 
Paragraph-level results with local context show improvements across all metrics compared to results without context, approaching chunk-level performance. This further verifies the supportive role of local context.
Local context improves CK-Recall from 19.20\% to 22.21\%, assisting the model to better understand the current paragraph. Moreover, global context can also yield some benefits, while not as significant as local context. Combining both local and global contexts, further improvements are seen in some metrics.

\vspace{-2mm}
\subsection{Results of Auxiliary Information}

We employ multiple methods to leverage auxiliary information. Table~\ref{tab:baseline} includes results of directly providing (Origin), retrieving the top-10 most relevant sentences (RAG), based on a summary (Summary), or list of keywords (Keywords) derived from LLMs. 

\textbf{Auxiliary information can enhance ASR error correction, while requiring more effective methods.}
Based on the GPT-4 model at chunk-level granularity, directly given auxiliary information (Origin) increases CK-Recall from 22.49\% to 43.70\%. 
The improvement of RAG is limited, indicating a need for better retrieval methods.
The Summary method improves objective metrics but shows limited CK-Recall gains, possibly due to missing keywords in the summary.
The Keywords method is most effective, further boosting recall to 45.42\%, and enhancing faithfulness scores while maintaining formality scores basically. 
However, improvements in CK-Recall stay below expectations, suggesting the limited benefits of current methods.
Effectively utilizing auxiliary information still needs further research.

\vspace{-2mm}
\section{Analysis}

\begin{table}[t!]
\begin{small}
    \centering
    \begin{tabular}{lccc}
    \toprule
    \textbf{\fontsize{8pt}{10pt}\selectfont{Model}} & \textbf{\fontsize{8pt}{10pt}\selectfont{\#Tokens}} & \textbf{\scriptsize{BLEURT$\uparrow$}} & \textbf{\fontsize{8pt}{10pt}\selectfont{Blank\%$\downarrow$}} \\
    
    \midrule
    GPT-4 (paragraph) & \phantom{0}340.50 & 63.32 & \phantom{0}0.00 \\
    \quad w/ local context & \phantom{0}654.47 & 67.11 & \phantom{0}0.00 \\
    GPT-4 (chunk) & 1515.00 & 62.15 & \phantom{0}1.60 \\
    \quad w/  auxiliary & 2006.86 & 62.73 & \phantom{0}1.28 \\
    GPT-4 (document) & 8140.42 & 10.64 & 76.10 \\

    \midrule
    QWen-14B (paragraph) & \phantom{0}340.50 & 59.87 & \phantom{0}0.00 \\
    \quad w/ local context & \phantom{0}654.47 & 55.64 & \phantom{0}0.00 \\
    QWen-14B (chunk) & 1515.00 & 30.14 & 36.56 \\
    \quad w/  auxiliary & 2006.86 & 22.35 & 48.33 \\
    QWen-14B (document) & 8140.42 & \phantom{0}2.87 & 91.58 \\
    
    \bottomrule
    \end{tabular}
    \vspace{-2mm}
    \caption{\small{The comparison of instruction-following abilities between GPT-4 and QWen-14B. Instructions are arranged by increasing difficulty levels and the number of input tokens. The fewer the blank paragraphs, the stronger the instruction-following ability.}}
    \label{tab:mismatch}
    \vspace{-2mm}
\end{small}
\end{table}

\begin{table}[t!]
\begin{small}
    \centering
    \begin{tabular}{lcccc}
    \toprule
    \multirow{2}{*}{\textbf{\fontsize{8pt}{10pt}\selectfont{Model}}} & \multicolumn{2}{c}{\textbf{\fontsize{8pt}{10pt}\selectfont{S-Faithful$\uparrow$}}} & \multicolumn{2}{c}{\textbf{\fontsize{8pt}{10pt}\selectfont{S-Formal$\uparrow$}}} \\
    \cline{2-5}
     & \textbf{\fontsize{8pt}{10pt}\selectfont{Total}} & \textbf{\fontsize{8pt}{10pt}\selectfont{Non-Blank}} & \textbf{\fontsize{8pt}{10pt}\selectfont{Total}} & \textbf{\fontsize{8pt}{10pt}\selectfont{Non-Blank}} \\
    
    \midrule
    GPT-4 & \textbf{7.32} & \textbf{7.39} & \textbf{8.20} & \textbf{8.29} \\
    QWen-Max & 6.00 & 6.50 & 7.55 & 8.21 \\
    QWen-14B & 3.72 & 4.99 & 6.04 & \textbf{8.29} \\
    LLaMA-8B & 6.35 & 6.53 & 7.33 & 7.56 \\
    
    \bottomrule
    \end{tabular}
    \vspace{-2mm}
    \caption{\small{Comparison of different LLMs among total paragraphs with non-blank paragraphs at the chunk-level granularity.}}
    \label{tab:blank}
    \vspace{-6mm}
\end{small}
\end{table}

\subsection{Instruction Following Capability}

We observe that \textbf{larger LLMs (e.g., GPT-4) show stronger instruction-following ability} compared to smaller LLMs (e.g., QWen-14B). 
We instruct the models to organize their responses by aligning each paragraph with the corresponding content using the paragraph index.
We refer to paragraphs without aligned paragraph indices as \textit{blank paragraphs}, which do not have corresponding target results. The rate of blank paragraphs can reflect a model's instruction-following capability. The fewer the blank paragraphs, the stronger the instruction-following ability. 
Thus we compute the rate of blank paragraphs as shown in Table~\ref{tab:mismatch}. At the chunk-level granularity, GPT-4 demonstrates robust alignment with scarcely any blank paragraphs (1.60\%), whereas QWen-14B exhibits more alignment errors (36.56\%).
Given the auxiliary information, as the number of input tokens increases, QWen-14B's performance further declines to 48.33\%, while GPT-4 remains virtually unaffected.

The differences in instruction-following ability also impact the model's performance on the CoS2W task. As shown in Table~\ref{tab:blank}, we compare the results among total paragraphs with non-blank paragraphs. 
Among non-blank paragraphs, QWen-14B's S-Formal performance is actually quite good (8.29). However, the overall performance (6.04) is relatively low due to the influence of blank paragraphs.

\vspace{-2mm}
\subsection{Discussion on LLM Evaluation}
\label{ssec:evaluator}

\begin{figure}[t!]
    \centering
    \includegraphics[width=\linewidth]{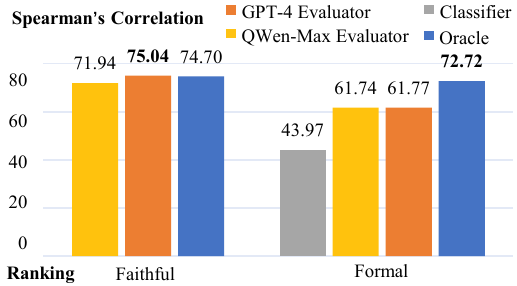}
    \vspace{-6mm}
    \caption{\small{The ranking correlation coefficients between different evaluation methods and human evaluation. Note that the ``Oracle'' represents the average scores of three annotators, and ``Classifier'' only reports results on English paragraphs.}
    }
    \label{fig:correlation}
    \vspace{-6mm}
\end{figure}

To measure the reliability of LLM evaluation, we compare the results with human evaluation. 
Consequently, we find that \textbf{the LLM Evaluator is reliable for both faithfulness and formality evaluation.}
Based on more than 200 sampled paragraphs, we engage three annotators to independently evaluate and rank the performance of GPT-4, QWen-Max, and QWen-14B in terms of faithfulness and formality. 
Then we calculate the Spearman's correlation coefficients~\cite{spearman1961proof} between the rankings given by the human evaluation and the LLM evaluation, as shown in Figure~\ref{fig:correlation}.

\textbf{Faithfulness Score}: The Spearman's correlation coefficient of the faithfulness rankings between the LLM Evaluator and human evaluation is 0.75, which is comparable to human performance. 
This demonstrates the reliability of the LLM Evaluator in faithfulness scoring.
Additionally, this indicates that the LLM performs well in determining whether the results faithfully preserve the original content.

\textbf{Formality Score}: 
As shown in Figure~\ref{fig:correlation}, the Spearman's correlation coefficient between the LLM evaluation and human evaluation is 0.61. 
Although there is a gap compared to human performance, there is also a strong correlation, which validates that the formality scores are reliable. 
Moreover, we compare the commonly used evaluation method. Using a RoBERTa-based classifier\footnote{\scriptsize\url{https://www.huggingface.co/s-nlp/roberta-base-formality-ranker}}, we classify English paragraphs as formal or informal and assign the formality probability as the formality score. The LLM evaluator outperforms the classifier in the correlation coefficient, confirming the reliability of LLM evaluation.

As Figure~\ref{fig:correlation} illustrates, GPT-4 Evaluator correlates more strongly with human evaluations than QWen-Max Evaluator, confirming its reliability. In addition, we evaluate GPT-4 results three times. The average standard deviations of faithfulness scores (0.19) and formality scores (0.11) across all samples confirm the robustness. Despite LLM updates rapidly, stable evaluation results can still be achieved with the same model version at a low temperature.

\vspace{-2mm}
\subsection{Distribution of Languages and Domains}

As indicated in Table~\ref{tab:language}, LLMs generally excel more in Chinese than English at the chunk level. There are more blank paragraphs in English (2.68\% for GPT-4) than in Chinese (0.83\%). 
This may be because LLMs prioritize fluency in English over Chinese, often merging paragraphs for better fluency at the expense of faithfulness.

As shown in Table~\ref{tab:domain}, \textbf{the podcast domain, with the longest auxiliary information, shows the most significant improvement in CK-Recall}, especially with the Keywords method. This indicates that plentiful auxiliary information aids ASR error correction improvement. Optimizing the use of long-text auxiliary information for further improvement is an area that warrants deeper exploration.

\begin{table}[t!]
\begin{small}
    \centering
    \begin{tabular}{lcccc}
    \toprule
    \multirow{2}{*}{\textbf{\fontsize{8pt}{10pt}\selectfont{Model}}} & \multicolumn{2}{c}{\textbf{\fontsize{8pt}{10pt}\selectfont{BLEURT$\uparrow$}}} & \multicolumn{2}{c}{\textbf{\fontsize{8pt}{10pt}\selectfont{S-Faithful$\uparrow$}}} \\
    \cline{2-5}
     & \textbf{\fontsize{8pt}{10pt}\selectfont{Chinese}} & \textbf{\fontsize{8pt}{10pt}\selectfont{English}} & \textbf{\fontsize{8pt}{10pt}\selectfont{Chinese}} & \textbf{\fontsize{8pt}{10pt}\selectfont{English}} \\
    
    \midrule
    GPT-4 & \textbf{66.95} & \textbf{55.52} & \textbf{7.83} & \textbf{6.80} \\
    QWen-Max & 57.67 & 40.79 & 6.99 & 5.00 \\
    QWen-14B & 36.73 & 21.05 & 4.77	& 2.63 \\
    LLaMA-8B & 36.55 & 46.03 & 7.33 & 5.36 \\
    
    \bottomrule
    \end{tabular}
    \vspace{-2mm}
    \caption{\small{The performance of LLMs across different languages.}}
    \label{tab:language}
    \vspace{-2mm}
\end{small}
\end{table}

\begin{table}[t!]
\begin{small}
    \centering
    \begin{tabular}{lccc}
    \toprule
    \multirow{2}{*}{\textbf{\fontsize{8pt}{10pt}\selectfont{Model}}} & \multicolumn{3}{c}{\textbf{\fontsize{8pt}{10pt}\selectfont{CK-Recall$\uparrow$}}} \\
    \cline{2-4}
     & \textbf{\fontsize{8pt}{10pt}\selectfont{Podcast}} & \textbf{\fontsize{8pt}{10pt}\selectfont{Meeting}} & \textbf{\fontsize{8pt}{10pt}\selectfont{Lecture}} \\
    
    \midrule
    GPT-4 (chunk) & 19.82 & 30.00 & 32.81 \\
    \quad w/  aux. (Origin) & 45.82 & 30.00 & 36.72 \\
    \quad w/  aux. (RAG) & 36.55 & 35.00 & \textbf{39.84} \\
    \quad w/  aux. (Summary) & 41.27 & 35.00 & 37.50 \\
    \quad w/  aux. (Keywords) & \textbf{47.27} & \textbf{45.00} & 37.50 \\
    
    \bottomrule
    \end{tabular}
    \vspace{-2mm}
    \caption{\small{The CK-Recall performance of GPT-4 with various auxiliary information methods across different domains.}}
    \label{tab:domain}
    \vspace{-6mm}
\end{small}
\end{table}

\vspace{-2mm}
\section{Conclusion}
To improve the readability of ASR transcripts, we propose the Contextualized Spoken-to-Written conversion (CoS2W) task, construct and make available the document-level and multi-domain Spoken2Written of ASR transcripts Benchmark (SWAB) dataset. 
Based on the SWAB dataset, we compare the performance of different LLMs at various granularity levels, verify the beneficial roles of contexts and auxiliary information, and find that it is worth further exploring how LLMs maintain faithfulness and utilize auxiliary information to enhance ASR Error Correction.
Compared with human evaluation, we find that the LLM Evaluator performs well in faithfulness and formality ranking with a good correlation.
In the future, we plan to expand the dataset scale to better support research on supervised training for this task. 

\vspace{-2mm}
\section{Limitations}

The study's limitation lies in the SWAB dataset having only a single target, which may inadequately reflect model performance through many automatic metrics. Future efforts will focus on offering diverse targets and developing more effective evaluation methods.
Note that the source data of SWAB are owned by the copyright holder. We only provide ASR transcripts and annotation targets, along with the corresponding links of audios and videos. The license of SWAB will be for research purposes only. 
Additionally, it is inevitable that evaluations are influenced by the biases of GPT-4, given that we obtain initial results through GPT-4 and also use it for experiments and evaluation.

\section{Ethics Statement}
\textbf{The SWAB dataset used in this research is strictly for academic and non-commercial purposes.} We implemented several measures to ensure compliance with ethical standards, as follows.
\begin{itemize}
\item{\textbf{Data Transparency and Anonymization.}} The Chinese meetings in the SWAB are sourced from the training set of the publicly available AliMeeting dataset, and the English meetings come from the publicly available AMI meeting dataset. For our collected podcasts and lectures, we only provide ASR transcripts after rigorous text anonymization processes and our annotations, to ensure transparency regarding the data sources and their usage while maintaining anonymity.
\item{\textbf{Data Access Compliance.}} To further ensure the ethical use of the dataset, we require researchers to contact us via email to confirm their compliance with ethical guidelines and the conditions outlined in our data usage declaration, before granting them access to the dataset. This procedure includes ensuring that they are aware of and adhere to the Personal Information Protection Law (PIPL) and any relevant legal frameworks regarding personal data usage.
\item{\textbf{Authorization.}} Any personal data should be used only with express authorization, ensuring lawful and fair processing in accordance with applicable laws.
\end{itemize}

\section{Acknowledgments}
We would like to express gratitude to the anonymous reviewers for their valuable feedback.

\bibliography{main}

\clearpage
\appendix


\section{Prompt of CoS2W Experiment}
This is the prompt template for the CoS2W experiment and data construction of the SWAB dataset, taking chunk-level meetings as an example.

\begin{tcolorbox}[colback=blue!2!white, colframe=blue!75, title=Prompt for CoS2W]
Suppose you are a professional recorder. Please polish the span of meeting in a written and formal style. The refined output should be free of noticeable ASR (Automatic Speech Recognition) errors and grammatical errors, faithfully preserve the original content, and exhibit a clear written style with excellent readability. Pleasure refer to the given meeting abstractive summarization. The given content is a transcript of the meeting using ASR, usually with a spoken or informal style, in the format [[line-id][transcription]. The requirements are as follows:

1. Please correct grammatical, spelling, and ASR errors by referring to the meeting abstractive summarization.

2. Please transfer the informal style such as redundant and repetitive into the formal style.

3. Please improve the fluency and coherence. You can adjust the sentence structure if appropriate.

4. The refined output must faithfully preserve the information of the given content. Do not add, delete or modify important information.

5. Please retain the original language for multi-lingual content.

6. The number of lines in the polished result must be consistent with the given content. The output format is [line-id][result].

The following is the given meeting abstractive summarization:

\{auxiliary\}

The following is the given context:

\{context\}

The following is the given span content of meeting:

\{content\}

The following is the refined output of the meeting span:
\end{tcolorbox}

\section{Prompt of Evaluation}
We initially attempt to evaluate faithfulness and formality scores within one prompt, but we find that formality scores are highly influenced by faithfulness problems. 
For samples with faithfulness problems, the LLM Evaluator will award a low score not only in faithfulness but also in formality, even when the formality is satisfied. 
Consequently, following commonly used methods~\cite{DBLP:journals/corr/abs-2304-13462}, we independently evaluate the faithfulness and formality scores of the target, which resulted in a good correlation with human evaluations as shown in Figure~\ref{fig:correlation}.

Here is the prompt for faithfulness evaluation of the CoS2W task, taking meetings as an example.

\begin{tcolorbox}[colback=green!5!white, colframe=green!80!black, title=Prompt for Faithfulness Evaluation]
Suppose you are a professional text editor, please evaluate the quality of the model's refined output. The original content is a paragraph from Automatic Speech Recognition (ASR), which may contain recognition errors, grammatical errors, and spoken/informal style expressions. The model's refined output should be free of noticeable ASR errors and grammatical errors, faithfully preserve the original content, and exhibit a clear written style with excellent readability. The human's refined output can be considered as one of the qualified polished result, which can be used as a reference for quality assessment. Please evaluate and score the model's refined output. Use a scoring range from 1 to 10 for each criterion, where 1 represents the worst performance and 10 represents the best performance. The return fields include:

1. ASR Error Correction Score: Check whether the text effectively corrects errors from automatic speech recognition. The higher the score, the more accurate the correction.

2. Faithful Score: Determine whether the text polishing results maintain the intent and content of the original speech. The higher the score, the more complete the preservation of the original meaning.

3. Evaluation Summary: Based on the scores above, briefly explain the reasons for the scores, provide an overall evaluation of the polishing quality, and suggest modifications.

The following is the original content:

\{source\}

The following is the human's refined output for reference:

\{human-target\}

The following is the model's refined output to evaluate:

\{model-target\}

Please provide scores for each criterion and an overall evaluation with json format:

\end{tcolorbox}

Given the source and the manually annotated target in the prompt, we use the LLM to evaluate the performance in terms of ASR error correction and faithfulness. 
However, ASR error correction scores are also significantly impacted by faithfulness problems. The Pearson correlation coefficient~\cite{pearson1895vii} between ASR error correction scores and faithfulness scores is as high as 0.96, which indicates nearly perfect correlations. 
For samples with faithfulness problems, the LLM Evaluator will award a low score not only in faithfulness but also in asr error correction scores, even if there are no ASR errors in the results.
Moreover, the correlation between ASR error correction scores and human evaluation is not strong. 
Therefore, we do not report the ASR error correction scores. Instead, we use the  Challenging Keyword Recall (CK-Recall) as the metric for the ASR error correction subtask.

Here is the prompt for formality evaluation of the CoS2W task, taking meetings as an example.

\begin{tcolorbox}[colback=green!5!white, colframe=green!80!black, title=Prompt for Formality Evaluation]
Suppose you are a professional text editor, please evaluate the quality of the model's refined output. The model's refined output should be free of grammatical errors, and exhibit a clear written style with excellent readability. Please evaluate and score the model's refined output. Use a scoring range from 1 to 10 for each criterion, where 1 represents the worst performance and 10 represents the best performance. The return fields include:

1. Grammar Score: Evaluate the grammatical accuracy of the text. The higher the score, the fewer grammatical errors there are.

2. Formal Score: Assess the written language style of the text, including the level of formality and suitability for written communication. The higher the score, the stronger the written style of the text.

3. Readability Score: Evaluate the fluency and overall readability of the text, including ease of understanding and natural expression. The higher the score, the better the readability of the text.

4. Evaluation Summary: Based on the scores above, briefly explain the reasons for the scores, provide an overall evaluation of the polishing quality, and suggest modifications.

The following is the model's refined output to evaluate:

\{target\}

Please provide scores for each criterion and an overall evaluation with json format:

\end{tcolorbox}

Given only the target result, we use the LLM to evaluate the performance in terms of grammar, formality, and readability. 
In terms of grammar, we find that nearly all LLM-generated results are free of grammatical errors, as confirmed by both LLM evaluation and human evaluation. Therefore, there is no need to report grammar scores. 
For readability scores, the Pearson correlation coefficient between readability scores and formality scores is as high as 0.97, which indicates nearly perfect correlations. 
Thus, the redundant scores have been omitted from the report.

\section{Prompt of Auxiliary Information}

Auxiliary information sometimes is comprehensive and detailed with long length, which poses challenges for LLMs in recognizing and utilizing useful information.
For example, in the Chinese podcast domain, the average length of auxiliary information is around 2,000 words as shown in Table~\ref{tab:dataset_big}. 
Therefore, we aim to condense the content to enable LLM to better leverage auxiliary information. Using the LLM, we design two methods: one for generating summaries and another for extracting key entities.

Here are the prompts for extracting summaries and keywords from auxiliary information.

\begin{tcolorbox}[colback=orange!5!white, colframe=orange!90!black, title=Prompt for Extracting Summaries]
Please summarize the provided podcast introduction into a concise English summary. You need to read the following content carefully and write a concise summary. The requirements are as follows:

1. Please accurately understand and recount the key information, ensuring all key points are covered.

2. Preserve all the key vocabulary from the original text, including but not limited to person, location, organization, date, proper nouns, professional terminology, and other named entities, especially those words that automatic speech recognition (ASR) may find challenging to identify correctly.

3. Narrate clearly, with logical coherence, concise language, and avoid redundancy.

The following is the provided podcast introduction:

\{auxiliary-information\}

The following is the summary:

\end{tcolorbox}

\begin{tcolorbox}[colback=orange!5!white, colframe=orange!90!black, title=Prompt for Extracting Keywords]
Based on the provided podcast introduction, please accurately identify and extract key entities of different categories. Make sure you read the given content thoroughly and identify and classify keywords from several categories, including but not limited to: person, location, organization, date, proper nouns, professional terms, podcast names, WeChat official account names, etc. Pay special attention to words that automatic speech recognition (ASR) may find challenging to identify correctly. For each category, create a list formatted as follows: the category as the key and the corresponding list of keywords as the value. Please be sure to identify accurately, ensuring the accuracy and completeness of the keywords.

The following is the provided podcast introduction:

\{auxiliary-information\}

The following is the json result of key entities of different categories:

\end{tcolorbox}

\begin{table}[t!]
\begin{small}
    \centering
    \begin{tabular}{lccc}
    \toprule
     & \textbf{\fontsize{8pt}{10pt}\selectfont{Ranking}} & \textbf{\fontsize{8pt}{10pt}\selectfont{Formula}} & \textbf{\fontsize{8pt}{10pt}\selectfont{Assigned Coefficient}}  \\
    
    \midrule
     & [1,1,1] vs [1,1,1] & 100.0 & 100.0 \\
     & [1,1,1] vs [1,1,2] & 62.50 & 50.00 \\
     & [1,1,1] vs [1,2,2] & 62.50 & 50.00 \\
     & [1,1,1] vs [1,2,3] & 50.00 & 0.00 \\
    
    \bottomrule
    \end{tabular}
    \vspace{-2mm}
    \caption{\small{The Spearman's correlation coefficient for identical rankings. We show both the formula and the assigned coefficient results.}}
    \label{tab:spearman}
    \vspace{-5mm}
\end{small}
\end{table}

\begin{table*}[t!]
    \centering
    \begin{tabular}{llcccccc}
    \toprule
    \textbf{\small{Data}} & \textbf{\small{\# Words}} & \textbf{\small{\# Sentences}} & \textbf{\small{\# Paragraphs}} & \textbf{\small{\# Turns}} & \textbf{\small{\# Speakers}} & \textbf{\small{\# Auxiliary}}  \\
    
    \midrule

    Chinese & 13071.10\footnotesize{±7624.45} & 556.07\footnotesize{±297.26} & 189.07\footnotesize{±138.93} & 169.03\footnotesize{±152.45} & 2.10\footnotesize{±1.03} & 753.63\footnotesize{±1059.36} \\
    \quad Podcast & 21610.10\footnotesize{±7878.83} & 683.10\footnotesize{±253.53} & 202.70\footnotesize{±82.16} & 191.70\footnotesize{±94.56} & 2.90\footnotesize{±0.74} & 2051.30\footnotesize{±898.29} \\
    \quad Meeting & 9926.10\footnotesize{±740.61} & 769.10\footnotesize{±151.40} & 324.80\footnotesize{±99.71} & 314.40\footnotesize{±102.57} & 2.40\footnotesize{±0.84} & 135.70\footnotesize{±14.24} \\
    \quad Lecture & 7677.10\footnotesize{±596.89} & 216.00\footnotesize{±25.82} & 39.70\footnotesize{±5.79} & 1.00\footnotesize{±0.00} & 1.00\footnotesize{±0.00} & 73.90\footnotesize{±16.41} \\

    \midrule

    English & 6039.97\footnotesize{±3673.92} & 426.43\footnotesize{±258.67} & 137.00\footnotesize{±94.08} & 81.73\footnotesize{±75.89} & 2.77\footnotesize{±1.74} & 165.43\footnotesize{±74.91} \\
    \quad Podcast & 10077.40\footnotesize{±3605.97} & 709.00\footnotesize{±251.84} & 232.00\footnotesize{±80.39} & 100.00\footnotesize{±51.07} & 3.30\footnotesize{±2.06} & 192.70\footnotesize{±78.33} \\
    \quad Meeting & 3860.50\footnotesize{±1610.35} & 324.40\footnotesize{±111.06} & 131.60\footnotesize{±54.34} & 144.20\footnotesize{±63.24} & 4.00\footnotesize{±0.00} & 212.60\footnotesize{±44.80} \\
    \quad Lecture & 4182.00\footnotesize{±819.03} & 245.90\footnotesize{±57.82} & 47.40\footnotesize{±11.31} & 1.00\footnotesize{±0.00} & 1.00\footnotesize{±0.00} & 91.00\footnotesize{±22.00} \\

    \midrule
    
    SWAB & 9555.53\footnotesize{±6912.06} & 491.25\footnotesize{±283.89} & 163.03\footnotesize{±120.53} & 125.38\footnotesize{±127.24} & 2.43\footnotesize{±1.45} & 459.53\footnotesize{±801.45} \\

    \bottomrule
    \end{tabular}
    \caption{\small{The statistic of Spoken2Written of ASR Transcripts Benchmark (SWAB) dataset. }}
    \label{tab:dataset_big}
    \vspace{-2mm}
\end{table*}

\section{Spearman's Correlation Coefficient}
\label{ap_spearman}

Table~\ref{tab:spearman} presents the calculation methods of Spearman's correlation coefficient on identical rankings.

Note that different LLMs actually perform quite well in terms of formality. As shown in Table 5, both larger and smaller LLMs achieve a high formality score on the non-blank paragraphs. 
Thus, annotators generally tend to view LLM outputs as well-written and formal and give them identical rankings (i.e., [1, 1, 1]). The proportion of identical formality rankings is quite high (63.93\%) compared to faithfulness ranking (6.06\%). 
However, the identical rankings are not defined in the Spearman correlation coefficient calculations. 
For distinct rankings $R(X_i)$ and $R(Y_i)$ with $n$ observations, it can be computed using the popular formula:
\begin{equation*}
    r_s = 1 - \frac{6\sum_{i=1}^{n}{(R(X_i) - R(Y_i))^2}}{n(n^2 - 1)}
\end{equation*}
where $R(X_i) - R(Y_i)$ is the difference between the two ranks of each observation.

To ensure these paragraphs are not overlooked, we have assigned the correlation coefficient based on the formula.

\section{Statistic of SWAB}
\label{ap_swab}

Table~\ref{tab:dataset_big} presents the statistics of the SWAB. 
It can be seen that both Chinese and English data contain sufficient paragraphs, providing ample support for the evaluation. 
Additionally, we have collected relevant auxiliary information. In particular, the podcast domain has a wealth of long text information.

\begin{table}[t!]
\begin{small}
    \centering
    \begin{tabular}{lcccc}
    \toprule
    \multirow{2}{*}{\textbf{Model}} & \textbf{Librispeech} & \textbf{Gigaspeech} & \multicolumn{2}{c}{\textbf{Wenetspeech}} \\
     & test\_clean & test & net & meeting \\
    
    \midrule
    Whisper & \textbf{4.00} & \textbf{13.50} & 8.87 & 11.25 \\
    Paraformer & 4.50 & 14.70 & \textbf{6.74} & \textbf{6.97} \\
    
    \bottomrule
    \end{tabular}
    \caption{\small{The Character Error Rate (CER) on ASR test sets. The smaller metric indicates better ASR recognition performance. Among them, Librispeech and Gigaspeech are the English test sets, while Wenetspeech is the Chinese test set.}}
    \label{tab:english_asr}
    \vspace{-6mm}
\end{small}
\end{table}

\section{ASR Performance}
The paper of the Paraformer~\cite{gao2022paraformer} only reported performance on a Mandarin test set. To illustrate its performance, we present a comparison between the Whisper and Paraformer model on English and Chinese test sets in Table~\ref{tab:english_asr}. 
Among them, Librispeech~\cite{DBLP:conf/icassp/PanayotovCPK15} and Gigaspeech~\cite{DBLP:conf/interspeech/ChenCWDZWSPTZJK21} are the English test sets,
while Wenetspeech~\cite{DBLP:conf/icassp/ZhangLGSYXXBCZW22} is the Chinese test set.
Both models have a similar number of parameters. However, Whisper benefits from over 680,000 hours of training data, while the Paraformer model utilizes only 50,000 hours of training data, giving Whisper a certain advantage in recognition performance.

\end{document}